\documentclass[sigconf]{acmart}

\AtBeginDocument{%
  \providecommand\BibTeX{{%
    \normalfont B\kern-0.5em{\scshape i\kern-0.25em b}\kern-0.8em\TeX}}}

\copyrightyear{2023}
\acmYear{2023}
\setcopyright{acmcopyright}\acmConference[WDC '23]{The 2nd Workshop on the security implications of Deepfakes and Cheapfakes}{July 10–14, 2023}{Melbourne, VIC, Australia}
\acmBooktitle{The 2nd Workshop on the security implications of Deepfakes and Cheapfakes (WDC ’23), July 10–14, 2023, Melbourne, VIC, Australia}
\acmPrice{0.00}
\acmDOI{10.1145/3595353.3595881}
\acmISBN{979-8-4007-0203-7/23/07.}

\begin{document}
	\fancyhead{}

\title{Discussion Paper: The Threat of Real Time Deepfakes}


\author{Guy Frankovits, Yisroel Mirsky}%
\email{guyfrank@post.bgu.ac.il, yisroel@bgu.ac.il}%
\affiliation{%
\institution{Ben-Gurion University\\Department of Software and Information Systems Engineering}
\city{Beersheba}
\country{Israel}}


\begin{abstract}
Generative deep learning models are able to create realistic audio and video. This technology has been used to impersonate the faces and voices of individuals. These ``deepfakes'' are being used to spread misinformation, enable scams, perform fraud, and blackmail the innocent. The technology continues to advance and today attackers have the ability to generate deepfakes in real-time. This new capability poses a significant threat to society as attackers begin to exploit the technology in advances social engineering attacks. In this paper, we discuss the implications of this emerging threat, identify the challenges with preventing these attacks and suggest a better direction for researching stronger defences.\end{abstract}


\begin{teaserfigure}
	\centering
  \includegraphics[width=0.7\textwidth]{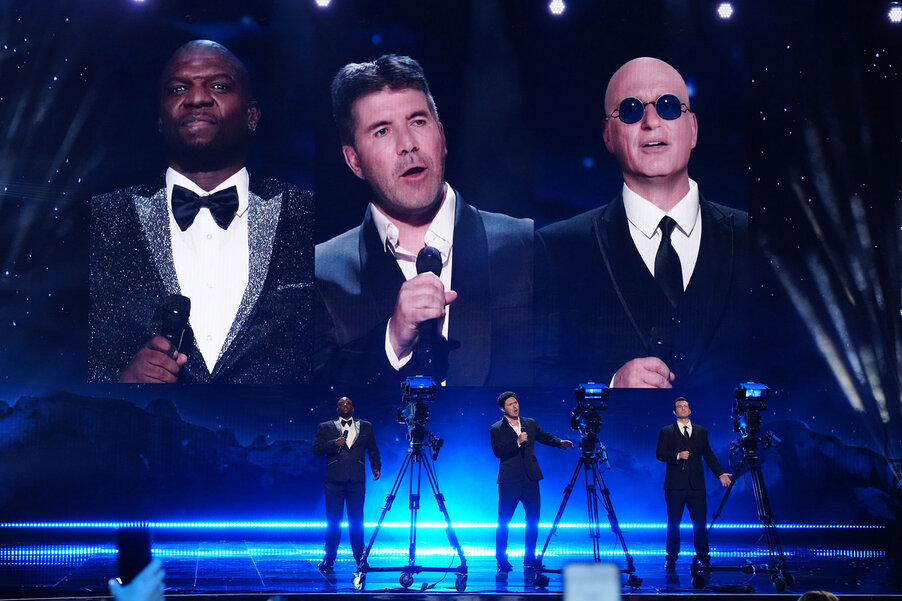}
  \caption{An example of a high quality real-time deepfake used in America's Got Talent \cite{AGT} created by Metaphysic \cite{metaphysic}. The ability to impersonate people in real-time poses a significant threat to society. This because voice and video calls can not longer be trusted based on content alone.}
  \label{fig:attacks}
  \vspace{1em}
\end{teaserfigure}

\maketitle

\section{Introduction}
A deepfake is synthetic media (e.g. video, audio) generated by a deep neural network, where the content looks authentic to a human \cite{mirsky2021creation}. The most common use of deepfake technology is to replace the identity of individuals in images and audio. In the year since its introduction in 2017, the technology has witnessed significant improvements in both the quality of the fake media it generates and the time it takes to generate them. With modern deepfake technology, it is possible to create content that looks authentic to humans. As a result, deepfakes have found numerous applications across various domains and industries. For example, the technology can be used to recreate a digital reanimation of the deceased \cite{Deepfakedeceased}, to enhance productivity in the film industry \cite{Deepfakepodactivity}, can be used in the education field \cite{Deepfakeeducation} and in many other areas where there is a need to create content in an easier and more flexible way.

Deepfake technology is highly accessible to even novice users \cite{perov2020deepfacelab, deepfakesweb, microsoftsVoiceColone, li2021starganv2}. This means that anyone can make convincing content of any individual. Malicious actors have noticed this capability and have used the technology for defamation, blackmail, misinformation, and social engineering attacks \cite{Reshapin92:online}. For example, in 2017 deepfake pornography surfaced on the Internet in 2017 raising serious concerns over the impact on women's privacy and dignity \cite{Deepfake76:online}. More recently, in March 2022, amidst the ongoing conflict between Russia and Ukraine, a deepfake video was circulated that depicted the Prime Minister of Ukraine instructing his troops to give up and stop fighting. This video had the potential to spread false information and undermine the credibility of the Ukrainian government and military \cite{Deepfake46:online}.

However, there is a new threat on the horizon: real-time deepfakes.\footnote{Examples of RT-DF tools:
\url{https://github.com/iperov/DeepFaceLive}\\
\url{https://github.com/alievk/avatarify-python}\\
\url{https://github.com/yl4579/StarGANv2-VC}\\
\url{https://github.com/CorentinJ/Real-Time-Voice-Cloning}}
Imagine a technician at a power plant who receives a phone call from his manager. The manager tells him to urgently change some configuration or provide him access to some resources. However, the technician is not speaking to his manager but rather an imposter. This scenario is possible with the latest deepfake technology which can generate a face or voice in real-time, enabling attackers to interact with their victims (an example of this technology is in Fig. \ref{fig:attacks}). This is a dangerous precedent because it gives attackers the ability to craft convincing false pretexts to manipulate their victims. This is because familiarity can be mistaken for authenticity. For example, in 2019 an attacker used the voice of a CEO to convince an associate to transfer \$243,000 in support of a made-up business emergency \cite{FraudstersVoice}. Another case was in 2021 where criminals pulled off a \$35 million dollar bank heist by tricking a banker by using a customer's voice \cite{Fraudste28:online}.

Real-time deepfakes (RT-DFs) also pose a threat to everyday people. For example, scammers can call up the elderly using their children's voices. This is feasible since some technologies only need a few seconds of the target's voice to clone their identity \cite{microsoftsVoiceColone}. In June 2022, the FBI released a warning that cyber criminals are using RT-DFs in job interviews in order to secure remote work positions and gain insider information. Then in August that year, cyber criminals attended Zoom meetings masquerading as the CEO of Binance \cite{Binancee98:online}.

\section{Discussion}
Deepfakes are not a new threat. Since their discovery in 2017,researchers have proposed a wide variety of ways to detect these attacks and mitigate their misuse \cite{almutairi2022review,lyu2020deepfake,yu2021survey,mirsky2021creation}. However, most of these defences are designed to work on offline media, not streaming media. Detecting and preventing real-time deepfake attacks is not as simple. Here are some aspects that make it hard to develop defences against these attacks:
\begin{description}
        \item[Practicality.] To protect users from fake calls, we can deploy detection models on the user's phone or laptop. However, this means that the models must be efficient. Moreover, they must also run in real-time to be able to monitor calls. These limitations are great challenges considering that most defences are based on deep learning which requires a considerable amount of resources to run. Moreover, attackers are cunning. If a detection model is only execute just before the call (e.g., as authentication), then the attacker could just switch between the real and deepfake content to evade detection.
        
        \item[Media Quality.] Many defences assume that deepfake generation models will leave semantic, stylistic, or forensic evidence in the generated content. However, voice and video calls often undergo significant data compression which degrades signals left in the media. This also means that attackers can intentionally increase the level of compression to cover up their defects without raising suspicion. 
        
        \item[Delivery.] Recently, the world has moved to a wider use of remote communication. With so many different ways once can communicate over a call or virtual meeting, it becomes difficult to develop a single solution that fits all scenarios. Moreover, some victims may still proceed with a flagged call, simply because they are worried that the system may be mistaken (imagine the caller being your boss, angry and insisting that you follow through with the call). 
\end{description}

\begin{figure*}
    \centering
    \includegraphics[width=.95\textwidth]{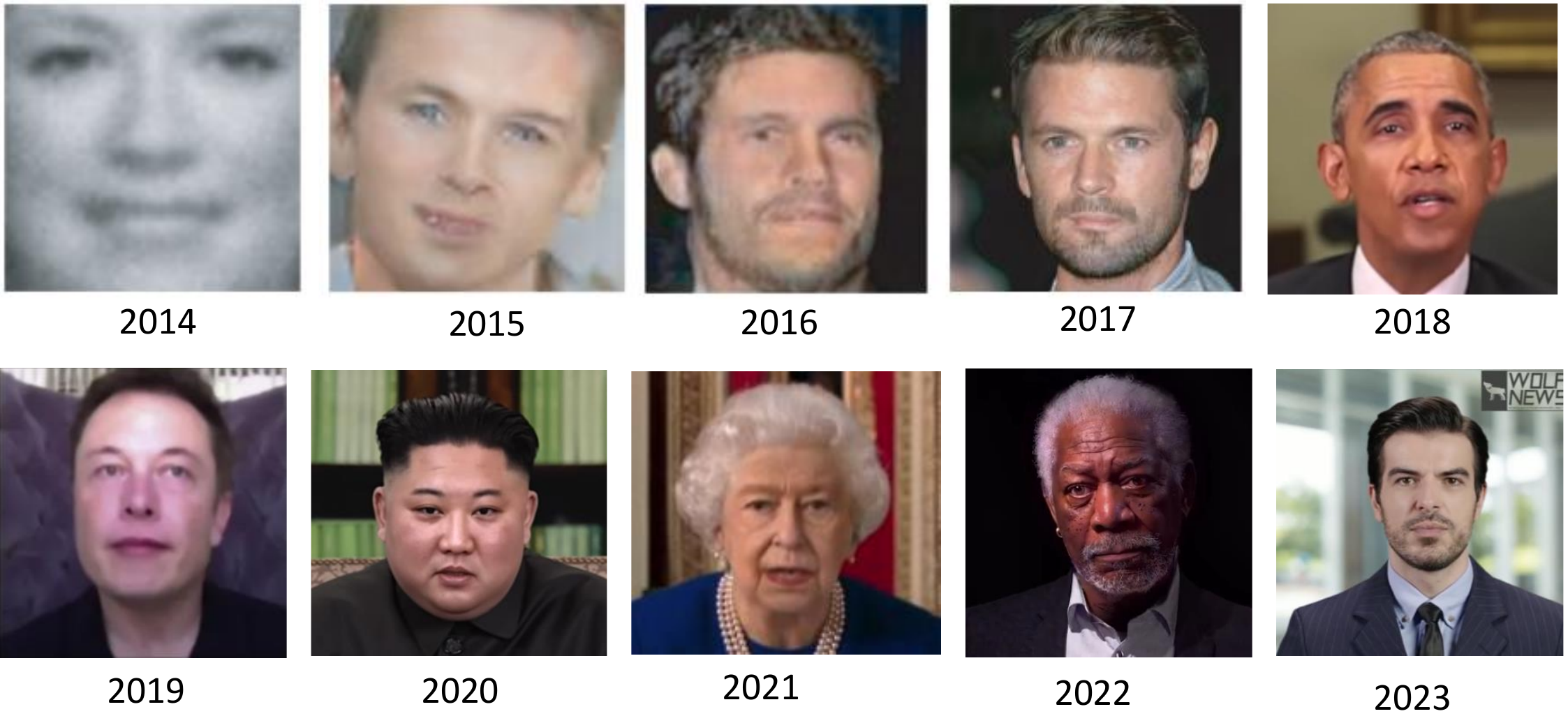}
    \vspace{-1em}
    \caption{A decade of deepfakes. The progression of facial deepfakes over the last ten years. 2014-2017 are images from GANs and such as StyleGAN. 2018 is from Buzzfeed. 2019 is a real-time deepfake that animates a single image. 2020-2023 are examples showing how deepfake quality is improving.}
    \label{fig:progression}
\end{figure*}

However, even if you could make a generic and practical detection model that can overcome compression, the model would still be looking for artifacts in the content itself. This is a losing battle. Looking at the advancements of generative AI over the last 10 years, it seems that `perfect' deepfakes are an inevitability (see Fig. \ref{fig:progression}). Therefore, we expect that \textit{passive} solutions that rely solely on artifacts, will be rendered ineffective in a matter of years. Moreover, there are other ways that attackers can hide evidence in the content. For example, adversarial machine learning can be used to make detection models classify deepfakes as real \cite{carlini2020evading}. Moreover, attackers can also remove specific artifacts. For example, models that search for blending boundaries can be mitigated by passing the final deepfake through a refiner such as a CycleGAN. Methods that look for biological signals \cite{hernandez2020deepfakeson} can be evaded by adding this ground truth to the model's \cite{li2020face} training. As a result, state-of-the-art defences do not provide long-term solutions.

To prepare for next-generation deepfakes, we cannot rely on passive methods of content analysis. In other words, we must seek solutions that expect that the media will reveal no forensic evidence of generative AI. There are at least two different defence strategies that accomplish this:
\begin{description}
    \item[Active] defences directly challenge the attacker. This gives the defender an advantage since he or she can now be proactive in his or her defence strategy. For example, in \cite{yasur2023deepfake} the authors propose using a new kind of CAPTHCA to detect deepfake calls. Callers are authenticated by performing a task that is hard for a deepfake model but easy for a human to perform (e.g., press on the cheek or turn around). This is essentially a new kind of Turing test where the challenge is on creating content as opposed to interpreting content.
    
    \item[Out of band] defences expose attacks by analyzing information surrounding the media. For example, by tracking the source of the media or by analysing the context surrounding the media. For example, a real-time deepfake caller could be revealed by simply determining the caller's origin or by verifying the status of the supposed caller (e.g., can we verify that the true individual is in a call right now?)
\end{description}
These approaches arguably provide better defences against the emerging threat real-time deepfakes. However, there is a lot of research which needs to be done to develop these defences. Therefore, we urge the research community to change paths and seek out more effective defences.

\section{Conclusion}
Real-time Deepfakes are an imminent threat and an urgent issue that must be dealt with. The technology provides malicious actors with the ability to perform powerful social engineering attacks. Although defences have been developed, many cannot be used to protect voice and video calls and many are likely to become obsolete as the quality of deepfakes improve. We encourage researchers to try to think of solutions that do not rely only on media content, thus giving the defenders a chance to keep ahead of the threat.
\section{acknowledgements}
This work was supported by the U.S.-Israel Energy Center managed by the Israel-U.S. Binational Industrial Research and Development (BIRD) Foundation and the Zuckerman STEM Leadership Program.



\bibliographystyle{ACM-Reference-Format}
\bibliography{bib}

\end{document}